\pdfoutput=1

\documentclass[11pt]{article}

\usepackage[]{acl}

\usepackage{times}
\usepackage{latexsym}

\usepackage[T1]{fontenc}

\usepackage[utf8]{inputenc}

\usepackage{microtype}

\usepackage{multirow}
\usepackage{graphicx}
\usepackage{enumitem}
\usepackage{amsmath,amssymb} 
\usepackage{fdsymbol}
\usepackage{booktabs}
\usepackage{amsfonts}
\usepackage{tabularx}               

\usepackage{xparse}
\NewDocumentCommand{\heng}
{ mO{} }{\textcolor{red}{\textsuperscript{\textit{Heng}}\textsf{\textbf{\small[#1]}}}}

\NewDocumentCommand{\cheng}
{ mO{} }{\textcolor{orange}{\textsuperscript{\textit{Cheng}}\textsf{\textbf{\small[#1]}}}}

\title{Improving Candidate Retrieval with Entity Profile Generation for Wikidata Entity Linking}

\author{Tuan Lai, Heng Ji, ChengXiang Zhai\\
	    University of Illinois at Urbana-Champaign \\
        \{tuanml2, hengji, czhai\}@illinois.edu\\
}

\begin{document}
\maketitle

\begin{abstract}

Entity linking (EL) is the task of linking entity mentions in a document to referent entities in a knowledge base (KB). Many previous studies focus on Wikipedia-derived KBs. There is little work on EL over Wikidata, even though it is the most extensive crowdsourced KB. The scale of Wikidata can open up many new real-world applications, but its massive number of entities also makes EL challenging. To effectively narrow down the search space, we propose a novel candidate retrieval paradigm based on entity profiling. Wikidata entities and their textual fields are first indexed into a text search engine (e.g., Elasticsearch). During inference, given a mention and its context, we use a sequence-to-sequence (seq2seq) model to generate the profile of the target entity, which consists of its title and description. We use the profile to query the indexed search engine to retrieve candidate entities. Our approach complements the traditional approach of using a Wikipedia anchor-text dictionary, enabling us to further design a highly effective hybrid method for candidate retrieval. Combined with a simple cross-attention reranker, our complete EL framework achieves state-of-the-art results on three Wikidata-based datasets and strong performance on TACKBP-2010\footnote{\;Our system is publicly available at \url{https://github.com/laituan245/EL-Dockers/}.}.
\end{abstract} 

\section{Introduction}

Entity linking (EL) is the task of mapping entity mentions in a document to standard referent entities in a target knowledge base (KB) \cite{SemTagAndSeeker,Cucerzan2007LargeScaleNE,Mihalcea2007WikifyLD,Milne2008LearningTL,ji2010overview,radhakrishnanetal2018elden,lai2021bert,jiangetal2021lnn}. EL systems have found applications in many tasks such as question answering \cite{lietal2020efficient}, knowledge base population \cite{dredzeetal2010entity}, information extraction \cite{li2020gaia,wen2021resin,lai2021joint}, and query interpretation \cite{queryinterpretation}. In general, the task is challenging because the same word or phrase can be used to refer to different entities. At the same time, the same entity can be referred to by different words or phrases.

Given the importance of EL, researchers have introduced a plethora of EL methods, ranging from using hand-crafted features \cite{ratinovetal2011local,Pan2015UnsupervisedEL} to using deep language models \cite{Agarwal2020EntityLV,decao2020autoregressive,bothaetal2020entity}. The vast majority of these studies have focused on linking mentions to Wikipedia or Wikipedia-derived KBs such as DBpedia \cite{Auer2007DBpediaAN} or YAGO \cite{Suchanek2007YagoAC}. As of November 2021, there are about 6.4 million articles in English Wikipedia. However, many entities are still missing from Wikipedia \cite{redi2020taxonomy}.

On the other hand, Wikidata, the most extensive general-interest KB, has much broader coverage than Wikipedia \cite{wikidata}. Wikidata contains more than 40 million entities with English titles, about seven times more than the number of articles in English Wikipedia. Every entity in Wikipedia has an equivalent entry in Wikidata, but not vice versa. The scale of Wikidata can open up many new real-world applications. When a disaster happens, many people rush to social media to share updates about the event \cite{Ashktorab2014TweedrMT}. Using an EL system to extract critical information (e.g., affected locations and donor agencies) can aid in monitoring the situation~\cite{Zhang2018}. However, many entities may not be well-known, and these entities are likely to be present in Wikidata than in Wikipedia \cite{Gei2017NECKArAN}. 

\begin{figure*}[!ht]
  \centering
  \includegraphics[width=\textwidth]{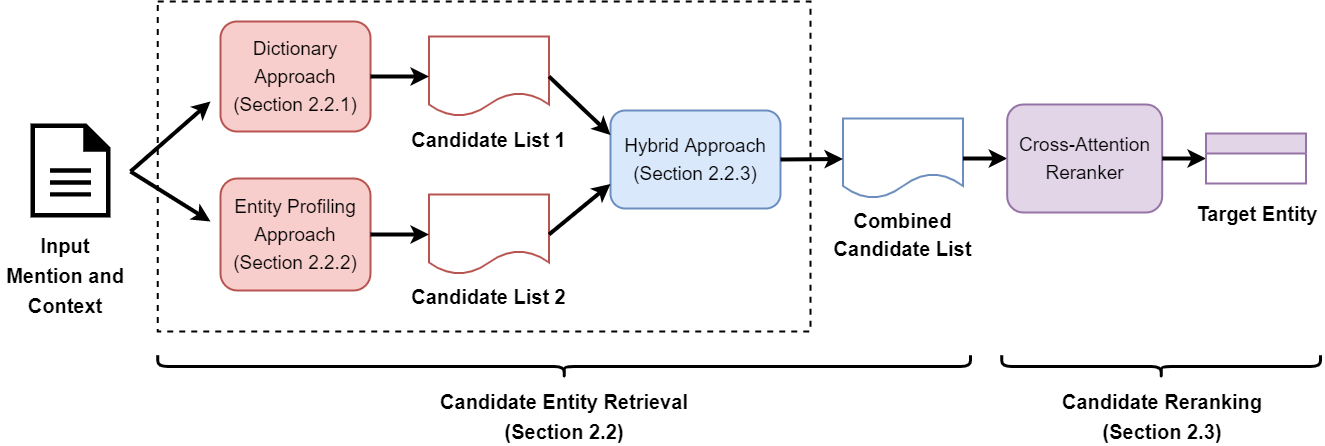}
  \caption{An overview of EPGEL, our entity linking framework.}
  \label{fig:framework_overview}
\end{figure*}

Despite the potential of Wikidata becoming a universal hub of real-world entities, there exists little in-depth research on EL over Wikidata \cite{mollersurvey}. The massive number of entities in Wikidata makes it challenging to find the correct entity for an input mention. Many previous EL methods for Wikipedia use a dictionary built from anchor texts to reduce the original search space to a small list of candidate entities \cite{Han2011CollectiveEL,Shen2015EntityLW,Phan2017NeuPLAS}. This dictionary-based approach is not directly applicable to Wikidata, since the description of each entity in Wikidata does not contain any anchor text.

In this work, we propose a novel candidate retrieval paradigm for Wikidata based on entity profile generation. Wikidata entities and their textual fields are first indexed into a text search engine (e.g., Elasticsearch). Given an entity mention and its context, we use a seq2seq model to generate the profile of the target entity, which consists of its title and description. The profile is then used to query the indexed search engine to retrieve candidate entities. Our technique is applicable to virtually any KB, not just Wikipedia or Wikidata. It also complements the dictionary-based approach, enabling us to further design an effective hybrid method for candidate retrieval. Combined with a simple cross-attention reranker, our complete EL framework achieves state-of-the-art (SOTA) results on three Wikidata-based datasets and strong performance on the standard TACKBP-2010 dataset.

In summary, our main contributions are: (1) a novel candidate retrieval paradigm based on entity profiling and (2) a new EL framework for Wikidata. Extensive experiments on four public datasets verify the effectiveness of our framework. We refer to our framework as \textbf{EPGEL}, which stands for \textbf{E}ntity \textbf{P}rofile \textbf{G}eneration for \textbf{E}ntity \textbf{L}inking.

\section{Methods}

\subsection{Overview}

\paragraph{Problem Formulation} Given a set of mentions $M = \{m_1, ..., m_N\}$ in a document and a knowledge base $\mathcal{E}$, the task is to find a mapping $M \rightarrow \mathcal{E}$ that links each mention to a correct entity in $\mathcal{E}$. We assume that entity mentions are already given, e.g., identified by some mention extraction module.

\paragraph{Entity Linking Framework} Figure \ref{fig:framework_overview} shows an overview of EPGEL. At a high level, similar to many previous methods \cite{Shen2015EntityLW}, EPGEL consists of two main stages: (1) candidate entity retrieval (2) candidate reranking. Given an entity mention, the role of the candidate retrieval module is to retrieve a small list of candidate entities (Sec. \ref{sec:candidate_entity_retrieval}). Our candidate retrieval approach is a combination of both the traditional dictionary-based approach (Sec. \ref{sec:dictionary_retrieval}) and our profiling-based approach (Sec. \ref{sec:profile_retrieval}). In the second stage, each candidate entity is reranked by a simple Transformer-based cross-attention reranker (Sec. \ref{sec:cross_attention_reranker}).


\begin{figure*}[!ht]
  \centering
  \includegraphics[width=\textwidth]{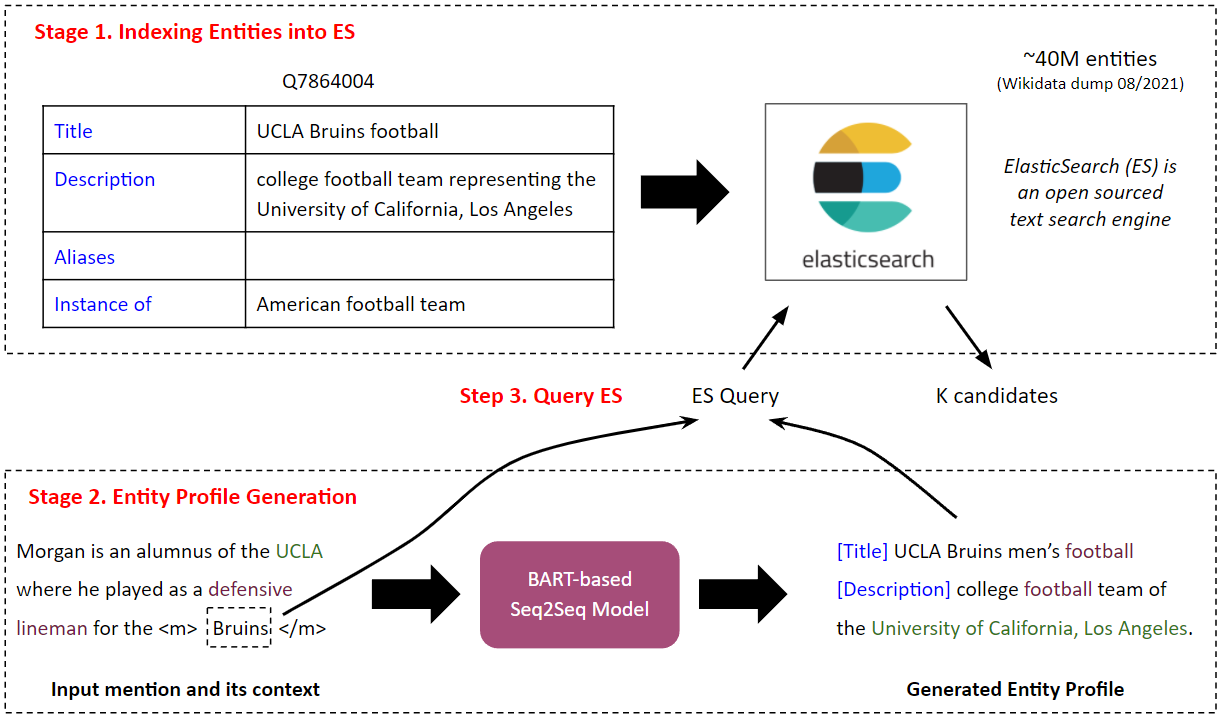}
  \caption{Candidate retrieval based on entity profiling.} 
  \label{fig:candidate_generation}
\end{figure*}

\subsection{Candidate Entity Retrieval} \label{sec:candidate_entity_retrieval}

\subsubsection{Dictionary-based Candidate Retrieval} \label{sec:dictionary_retrieval}
\paragraph{Overview} Dictionary-based techniques are the dominant approaches to candidate retrieval of many previous Wikipedia EL systems \cite{guoetal2013link,Ling2015DesignCF,fang2020high}. The basic idea is to estimate the mention-to-entity prior probability $\hat{p}(e|m)$. For example, both the technology company Amazon and the Amazon river could be referred to by ``Amazon''. However, when people mention ``Amazon'', it is more likely that they mean the company rather than the river.

\paragraph{Prior Estimation} The anchor texts in Wikipedia are frequently used for estimating the prior probability:
\begin{equation} \label{eq:prior_prob}
    \hat{p}(e|m) = \frac{\text{count}(m, e)}{\text{count}(m)} 
\end{equation}
where $\text{count}(m)$ is the total number of anchor texts having the entity mention $m$ as the surface form in Wikipedia; $\text{count}(m, e)$ denotes the number of anchor texts with the surface form $m$ pointing to the entity $e$. Even though this approach is highly effective for EL over Wikipedia \cite{ganeahofmann2017deep}, it is not directly applicable to Wikidata. A dictionary built from Wikipedia anchor texts will never return entities that are in Wikidata but not in Wikipedia. Furthermore, in Wikidata, the textual description of each entity is typically short and does not contain any anchor text. Therefore, it is not possible to build a dictionary specifically for Wikidata using the same approach. Below, we propose a new approach that is applicable to Wikidata.

\subsubsection{Entity Profiling for Candidate Retrieval} \label{sec:profile_retrieval}
\paragraph{Overview} We propose a more general paradigm for candidate retrieval (Figure \ref{fig:candidate_generation}). We first index all useful entities from Wikidata into Elasticsearch (ES), an open-source text search engine. During inference, given an entity mention and its context, we use a sequence-to-sequence (seq2seq) model to generate the profile of the target entity. We then use the original mention and the generated profile as the basis for formulating the ES query. This candidate retrieval approach based on entity profiling is applicable to virtually any KB. At the very least, each entity in a KB typically has a textual title.

\paragraph{Entity Profile Generation Model} A straightforward approach to query ES is to directly use the literal string of the input mention \cite{Sakor2020Falcon2A,kannanravietal2021cholan}. However, without any contextual information, the literal mention text is not informative and discriminative enough. In the example shown in Figure \ref{fig:candidate_generation}, one can simply ask ES to search for entities whose \textit{title} field or \textit{aliases} field contains the word ``Bruins''. However, there is an ice hockey team based in Boston named ``Bruins'' (Q194121), and there is also a college basketball team with the same name (Q3615392). Neither of these entities is the correct target entity (a football team of UCLA). In the input context, the phrase ``defensive lineman'' implies that the mention refers to a football team. Also, as UCLA is a common acronym abbreviating the University of California, Los Angeles, a well-trained generation model can generate a description that closely resembles the target entity's actual description (Figure \ref{fig:candidate_generation}). 

To this end, we train a conditional generation model for generating the profile of the target entity, where the condition is the mention and its context:
\[\text{[s]\;$\text{ctx}_{\text{left}}$\;[m]\;\textit{mention}\;[/m]\;$\text{ctx}_{\text{right}}$\;[/s]}\]
Here, \textit{mention}, $\text{ctx}_{\text{left}}$, $\text{ctx}_{\text{right}}$ are the tokens of the entity mention, context before and after the mention respectively. $[m]$ and $[/m]$ are used to separate the original mention from its context. $[s]$ and $[/s]$ are special tokens denoting the start and the end of the entire concatenated input, respectively. The target output is a concatenation of the target entity's title and its description (Figure \ref{fig:candidate_generation}).

Our conditional generation model is an encoder-decoder language model (e.g., BART \cite{lewisetal2020bart} and T5 \cite{Raffel2020ExploringTL}). The generation process models the conditional probability of selecting a new token given the previous tokens and the input to the encoder.
\begin{equation}
p(\textbf{Y}_{1:n} | c) = \prod_{i=1}^n p(\textbf{Y}_i\,|\,\textbf{Y}_{<i}, c)
\end{equation}
where $\textbf{Y}_{1:n}$ denotes the target output sequence and $c$ denotes the condition (i.e., the input mention and its context). 

\paragraph{Elasticsearch Query Construction} We directly use the original mention and the generated profile as the basis for formulating the ES query. We ask ES to score each entity based on the following criteria: (1) The similarity between the \textit{title} and \textit{alias} fields and the literal mention text. (2) The similarity between the \textit{title} and \textit{alias} fields and the generated title (3) The similarity between the \textit{description} field and the generated description. More details are in the appendix due to space constraints.

\subsubsection{Hybrid Approach to Candidate Retrieval} \label{sec:hybrid_cg}
\paragraph{Overview} Our main goal is to perform EL to Wikidata. However, a source document often contains entity mentions that can be linked to Wikipedia since Wikipedia still covers many fields and areas of interest. In addition, every entity in Wikipedia can be automatically mapped to an equivalent entity in Wikidata. As such, we propose a hybrid approach that combines both the dictionary-based technique (Section \ref{sec:dictionary_retrieval}) and our profiling-based retrieval technique (Section \ref{sec:profile_retrieval}). We first combine the lists produced by these two methods into one single candidate list. We then use a Gradient Boosted Tree (GBT) model \cite{Friedman2001GreedyFA} to assign a score to every candidate. Finally, the combined list is sorted based on the candidates' computed scores.

\paragraph{Combining Candidate Lists} For a mention $m$, let $C_d(m)$ be the set of candidates retrieved by a Wikipedia-based dictionary. Let $C_e(m)$ be the set of candidates retrieved by querying ES using generated entity profiles. We train a GBT model that assigns a score to every candidate in the combined set $C_d(m) \cup C_e(m)$. We use two groups of features: string-based and ranking-based features.

For string-based features, we use several similarity metrics: (1) Levenshtein ratios \cite{Levenshtein1965BinaryCC}, Jaro–Winkler distances \cite{Jaro1989AdvancesIR}, and numbers of common words between the mention's surface form and the candidate entity's name and aliases (2) Numbers of common words between the mention's context and the entity's name and aliases (3) Numbers of common words between the mention's surface form and context and the entity's description and category.

We also use features that indicate the initial rankings of a candidate entity. For $C_d(m)$, each candidate is initially ranked by its corresponding prior probability (Eq. \ref{eq:prior_prob}). For $C_e(m)$, each candidate is automatically assigned a score by ES. For a candidate $c$, let $r_d(c)$ indicate its rank in $C_d(m)$ (if $c \notin C_d(m)$ then $r_d(c) = \infty$). Similarly, let $r_e(c)$ indicate the rank of $c$ in $C_e(m)$. The features to be fed to GBT are:
\begin{equation}
\begin{split}
a_d(c) &= \begin{cases}
1/r_d(c), & \text{if $c \in C_d(m)$}\\
0, & \text{Otherwise}
\end{cases}\\
a_e(c) &=  \begin{cases}
1/r_e(c), & \text{if $c \in C_e(m)$}\\
0, & \text{Otherwise}
\end{cases}\\
\end{split}
\end{equation}


\subsection{Cross-Attention Reranker} \label{sec:cross_attention_reranker}

\begin{figure}[!t]
  \centering
  \includegraphics[width=\linewidth]{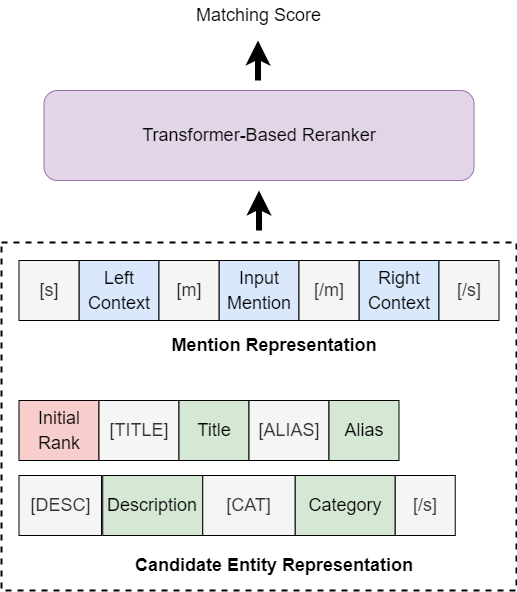}
  \caption{An illustration of the cross-attention reranker.}
  \label{fig:reranker}
\end{figure}

\paragraph{Overview} We model the reranking problem as a binary classification problem and fine-tune a basic Transformer-based reranker for the task (Figure \ref{fig:reranker}).

\paragraph{Input Representations} The input to the reranker is the concatenation of the mention representation and the candidate entity representation (Figure \ref{fig:reranker}). The mention representation is similar to the one described in Section \ref{sec:profile_retrieval}. Each entity's representation consists of its initial rank (Section \ref{sec:hybrid_cg}), title, alias, description, and category. To denote the initial rank, we define new tokens in the Transformer's vocabulary. For example, $[\text{rank1}]$ represents rank 1, $[\text{rank2}]$ indicates rank 2, and so on. If an entity has multiple aliases, we select the one with the highest string similarity to the input mention. The special tokens $[\text{TITLE}]$, $[\text{ALIAS}]$, $[\text{DESC}]$, and $[\text{CAT}]$ are used to indicate the locations of the entity's title, alias, description, and category (respectively). If any fields are missing, we simply exclude the missing fields and their corresponding special tokens from the entity representation.

\paragraph{Cross-Attention Reranker} Given a mention $m$ and a candidate entity $e$, the reranker computes a matching score $s_{m,e}$ indicating their relevance. The reranker consists of a Transformer-based encoder and a feedforward network:
\begin{equation}
\begin{split}
  \textbf{h}_{m,e} &= \text{reduce}(T_{\text{cross}}(\tau_{m,e})) \\[0.5em]
  s_{m,e} &= \text{FFNN}_{s}(\textbf{h}_{m,e})
\end{split}
\end{equation}
where $\tau_{m,e}$ is the concatenation of the mention representation and the entity representation. $T_{\text{cross}}$ is a Transformer encoder \cite{devlinetal2019bert,Liu2019RoBERTaAR}, and $\text{reduce}(.)$ is a function that returns the final hidden state of the Transformer that corresponds to the first token (i.e., the $[\text{s}]$ token). $\text{FFNN}_{\text{s}}$ is a feedforward network. By taking $\tau_{m,e}$ as input, the Transformer encoder $T_{\text{cross}}$ can have deep cross-attention between the mention's context and the entity's information from the KB.

In practice, a mention may not have any corresponding entity in the target KB. For predicting unlinkable mentions, we employ a simple thresholding method. If the score $s_{m, e_{top}}$ of the top-ranked candidate entity $e_{top}$ is smaller than a threshold, we predict the mention $m$ as unlinkable.
\section{Experiments}
\subsection{Data and Experiments Setup} \label{sec:data_and_exp}

\renewcommand{\arraystretch}{1.2}
\begin{table*}[ht!]
\centering
\resizebox{\textwidth}{!}{%
\begin{tabular}{lccc|ccc|ccc|ccc}
\hline
\multicolumn{1}{c}{\multirow{2}{*}{Methods}} & \multicolumn{3}{c|}{RSS-500 (test)} & \multicolumn{3}{c|}{ISTEX-1000 (test)} & \multicolumn{3}{c|}{TweekiGold (test)} & \multicolumn{3}{c}{WikipediaEL (dev)} \\ \cline{2-13} 
\multicolumn{1}{c}{} & R@1 & \multicolumn{1}{c}{R@25} & R@50 & R@1 & R@25 & R@50 & R@1 & R@25 & R@50 & R@1 & R@25 & R@50 \\ \hline
Simple Query & 41.06 & 72.19 & 74.17 & 36.42 & 79.10 & 90.15 & 31.02 & 73.96 & 82.52 & 51.19 & 81.85 & 85.86 \\
Wikipedia Dictionary & 59.60 & 74.83 & 76.82 & 84.93 & 91.49 & 91.49 & 70.60 & 88.08 & 88.77 & 85.11 & 93.60 & 93.95 \\\hline
Profiling-based Query &  &  &  &  &  &  &  &  &  &  &  & \\
$\mdblkdiamond$ Title & 49.00 & 73.51 & 76.82 & 43.28 & 82.69 & 93.28 & 39.81 & 79.86 & 87.03 & 54.77 & 88.19 & 92.13\\
$\mdblkdiamond$ Title + Desc & 60.26 & 73.51 & 75.50 & 87.61 & 97.31 & 98.06 &   71.30 & 88.77 & 91.55 & 80.87 & 94.26 & 95.03\\\hline
Hybrid Approach & \textbf{66.89} & \textbf{85.43} & \textbf{86.09} & 
\textbf{91.34} & \textbf{98.51} & \textbf{98.66} & \textbf{74.54} & \textbf{95.14}  & \textbf{95.60} & \textbf{90.25} & \textbf{98.95} & \textbf{99.23}
\end{tabular}%
}
\caption{Overall candidate retrieval results. Recall scores (\%) are shown.}
\label{tab:overall_cg_results}
\end{table*}

\paragraph{Target Knowledge Base} In this work, we downloaded the complete Wikidata dump dated August 2021. Wikidata currently contains over 95 million items. However, many of these items are noisy or correspond to Wikimedia-internal administrative entities (i.e., not entities we want to retain). Therefore, we apply several heuristics to filter out unhelpful Wikidata items\footnote{\;More details are in the appendix.}. At the end, our final knowledge base contains 40,239,259 entities with English titles, substantially more than any other task settings we have found. We use this KB as the target KB for every EL experiment we conduct.

\paragraph{Evaluation Datasets (Wikidata)} We use three manually annotated English datasets for evaluating EL over Wikidata: \textbf{RSS-500} \cite{Rder2014NA}, \textbf{ISTEX-1000} \cite{Delpeuch2020OpenTapiocaLE}, and \textbf{TweekiGold} \cite{harandizadehsingh2020tweeki}. More details of these datasets are in the appendix. Some previous studies on EL over Wikidata also use other datasets such as LC-QuAD 2.0 \cite{Dubey2019LCQuAD2A} and  T-REx \cite{ElSahar2018TRExAL}. However, these datasets were created semi-automatically or automatically instead of manually, thus less reliable.

\paragraph{Training Data} We use Wikipedia anchor texts and their corresponding Wikidata entities as the supervision signals. We create a training set of 6 million paragraphs and a validation set of 1000 paragraphs. We refer to this dataset as \textbf{WikipediaEL}. We train our models (i.e., the generation model and the reranker) using this dataset. We do not fine-tune our models on any of the evaluation datasets.

\paragraph{Baselines} For comparison, we choose a set of systems that were previously evaluated on the same evaluation datasets: AIDA \cite{hoffartetal2011robust}, Babelfy \cite{babelfly}, End-to-End \cite{kolitsasetal2018end}, OpenTapioca \cite{Delpeuch2020OpenTapiocaLE}, Tweeki \cite{harandizadehsingh2020tweeki}, and KG Context \cite{mulang2020evaluating}.

We also compare our approach to BLINK \cite{wuetal2020scalable} and GENRE \cite{decao2020autoregressive}, SOTA methods for EL over Wikipedia or Wikipedia-derived KBs. We evaluated these methods by using their public code and model checkpoints. We implemented a converter to map each returned entity to its corresponding Wikidata entry.

CHOLAN \cite{kannanravietal2021cholan} is a related study, but its open-sourced code lacks running instructions\footnote{\,\url{https://tinyurl.com/el-cholan}}. Furthermore, the authors have not fully disclosed the splits of the dataset they used for evaluating EL over Wikidata. As a result, we did not directly compare CHOLAN and EPGEL.

\paragraph{Hyperparameters} Our generation model is initialized with the BART model (bart-base) \cite{Lewis2020BARTDS}. For the reranker, we use RoBERTa (roberta-base) as the Transformer encoder \cite{Liu2019RoBERTaAR}. The maximum numbers of candidates are set to be 100, 100, and 50 for the dictionary-based, profiling-based, and hybrid approaches (respectively). More details are in the appendix.



\renewcommand{\arraystretch}{1.2}
\begin{table*}[!ht]
\centering
\resizebox{\textwidth}{!}{%
\begin{tabular}{lc|c|c|c}
\hline
\multicolumn{1}{c}{\multirow{2}{*}{Methods}} & \multirow{2}{*}{RSS-500 (test)} & \multirow{2}{*}{ISTEX-1000 (test)} & \multicolumn{1}{c|}{\multirow{2}{*}{TweekiGold (test)}} & \multicolumn{1}{c}{\multirow{2}{*}{WikipediaEL (dev)}} \\
\multicolumn{1}{c}{} &  &  & \multicolumn{1}{c|}{} & \multicolumn{1}{c}{} \\ \hline
EPGEL & \textbf{76.4} & \textbf{92.7} & \textbf{69.3} & \textbf{92.3} \\ \cline{1-5}
\textit{Effects of Candidate Retrieval Strategy} & & & & \\
$\mdblkdiamond$ Simple Query & 66.4 & 87.6 & 66.0 & 81.9 \\
$\mdblkdiamond$ Wikipedia Dictionary & 71.2 & 91.6 & 68.8 & 89.8 \\
$\mdblkdiamond$ Profiling-Based Query [Title + Desc] & 68.4 & 92.6 & 69.1 & 88.4 \\ \hline
\textit{Previous Methods} & & & &\\
GENRE ${}^{\star}$ \cite{decao2020autoregressive} & 68.2 & 88.4 & 62.4 & 86.3 \\
BLINK ${}^{\star}$ \cite{wuetal2020scalable} & 73.5 & 88.5 & 65.9 & 90.5 \\
KG Context ${}^\dagger$ \cite{mulang2020evaluating} & - & 92.6 & - & - \\
Tweeki \cite{harandizadehsingh2020tweeki} & - & - & 65.0 & - \\
OpenTapioca \cite{Delpeuch2020OpenTapiocaLE} & 46.5 & 91.6 & 29.1 & - \\
End-to-End \cite{kolitsasetal2018end} & - & - & 49.4 & - \\
Babelfy \cite{babelfly} & 58.1 & 64.0 & 25.1 & - \\
AIDA \cite{hoffartetal2011robust} & 56.1 & 50.4 & 38.5 & - 
\end{tabular}%
}
\caption{Overall entity linking results. \textit{InKB} micro F1 scores (\%) are shown. The symbol ``-'' denotes results not reported in previous papers. The symbol ``$\star$'' indicates systems that we evaluated by ourselves using their public code and model checkpoints. ${}^\dagger$ KG Context is reported to have an F1 score of 92.6 on ISTEX-1000 \cite{mulang2020evaluating}. However, the work uses a simplified setting where each mention's candidate pool is assumed to consist of the correct entity and only one negative entity. This setting is much easier and less practical than our setting.}
\label{tab:overall_el_results}
\end{table*}
\subsection{Evaluation of Candidate Entity Retrieval} \label{sec:cg_retrieval_evaluation}

Table \ref{tab:overall_cg_results} compares the performance of various candidate retrieval approaches. [Simple Query] refers to querying ES using only the literal string of the input mention. This approach is quite similar to what is done in several previous studies on EL over Wikidata \cite{Sakor2020Falcon2A,kannanravietal2021cholan}. 
As the target KB is huge, many entities have the same titles or aliases. Naively using only the surface form of the mention is not sufficient. 

The performance of using a Wikipedia dictionary (Section \ref{sec:dictionary_retrieval}) is much better than that of [Simple Query]. Although the dictionary-based approach also does not consider the context of the input mention, it computes the conditional probabilities using all anchor texts in the entire Wikipedia. In addition, most target entities in the evaluation datasets can still be found in Wikipedia. As such, this approach still performs reasonably well overall. However, note that for mentions whose linked entities are in Wikidata but not in Wikipedia, the recall score of the Wikipedia dictionary will always be 0.

For our profiling-based approach (Section \ref{sec:profile_retrieval}), we experiment with two variants: (1) The entity profile is only the generated title (2) The entity profile consists of the generated title and the generated description. The latter achieves much better performance. It also achieves comparable or better scores than the Wikipedia dictionary most of the time.

Finally, we see that our profiling-based approach complements the dictionary-based approach. Our hybrid technique (Section \ref{sec:hybrid_cg}) is highly effective, outperforming all other methods.
\subsection{Overall Entity Linking Results} \label{sec:el_results_section}
Table \ref{tab:overall_el_results} shows the overall entity linking results. Our complete framework (i.e., EPGEL) uses the hybrid candidate retrieval approach (Section \ref{sec:hybrid_cg}) and the cross-attention reranker (Section \ref{sec:cross_attention_reranker}). EPGEL outperforms a variety of SOTA techniques across all datasets. For example, EPGEL achieves better results than GENRE \cite{decao2020autoregressive} on the tested datasets. GENRE is an autoregressive system that directly retrieves entities by generating the entity names conditioned on the context. In theory, GENRE does not require a candidate retrieval step to work. However, as detailed in the original paper \cite{decao2020autoregressive}, GENRE achieves the best performance when high-quality candidate lists are available. Therefore, having an effective candidate retrieval method can still be helpful even during this era of large language models.

Table \ref{tab:overall_el_results} also shows the results of using different candidate retrieval strategies. There is a positive correlation between the candidate retrieval performance and the final EL performance. This is expected, as the recall from the candidate retrieval step provides an upper bound on the entire EL framework's recall. Also, even if EPGEL uses only the profiling-based approach (without relying on the Wikipedia dictionary), it can still achieve good results compared to the baselines.
\renewcommand{\arraystretch}{1.1}
\begin{table}[!t]
\centering
\resizebox{\linewidth}{!}{%
\begin{tabular}{lc}
\hline
\multicolumn{1}{c}{Methods} & P@1 \\ \hline
Neural Cross-Lingual EL \cite{Sil2018NeuralCE} & 87.4\\
DeepType \cite{Raiman2018DeepTypeME} & 90.9\\
Neural Collective EL \cite{Cao2018NeuralCE} & 91.0\\
DEER \cite{Gillick2019LearningDR} &  87.0 \\
RELIC \cite{Ling2020LearningCE} & 89.8 \\
Attribute-sep. \cite{vyasballesteros2021linking} & 84.9 \\ \hline
EPGEL & 90.9 \\
\end{tabular}%
}
\caption{In-KB accuracy scores (\%) of different models on TACKBP-2010. Note that our Wikidata-based target KB is much larger than the ones used by previous studies (e.g., the TAC Reference KB).}
\label{tab:tackbp_results}
\end{table}


\subsection{Results on TACKBP-2010}
Even though our focus is EL over Wikidata, we also use the TACKBP-2010 dataset \cite{ji2010overview} for evaluation since it is a standard dataset used by many previous studies. There are 1,020 annotated mention/entity pairs in total for evaluation. All the entities are from the TAC Reference KB, containing only 818,741 entities. To evaluate EPGEL, we use our large-scale Wikidata-based KB as the target KB. Also, we do not fine-tune EPGEL on the training set of TACKBP-2010. Overall, the performance of EPGEL is comparable to previous state-of-the-art systems (Table \ref{tab:tackbp_results}), even though EPGEL needs to map mentions to entities in a large-scale KB.
\subsection{Qualitative Analysis}

\renewcommand{\arraystretch}{1.75}
\begin{table*}[!ht]
\centering
\resizebox{\textwidth}{!}{%
\begin{tabular}{|m{0.55\textwidth}|m{0.4\textwidth}|c|}
\hline
Input Mention & Generated Profile & Target Entity \\ \hline
... They had an only son, Arthur, a British Army officer who played a leading role in the 1914 \textcolor{red}{Christmas truce}. & \textcolor{blue}{[Title]} Christmas truce $\vert$ \textcolor{blue}{[Description]} unofficial cease fire in Western Front during World War I & \href{https://www.wikidata.org/wiki/Q163730}{Q163730} \\ \hline
... and as a member of the National Football League. It also marked the 14th season under leadership of general manager \textcolor{red}{Kevin Colbert} and the seventh under head ... & \textcolor{blue}{[Label]} Kevin Colbert $\vert$ \textcolor{blue}{[Description]} American football executive & \href{https://www.wikidata.org/wiki/Q6396037}{Q6396037} \\ \hline
... \textcolor{red}{Baltimore} beat Josh Beckett and the Red Sox 7-1 Tuesday night ... & \textcolor{blue}{[Title]} Baltimore $\vert$ \textcolor{blue}{[Description]} Independent city in Maryland, United States & \href{https://www.wikidata.org/wiki/Q650816}{Q650816} \\ \hline
\end{tabular}%
}
\caption{Example outputs from our conditional generation model.}
\label{tab:qualitative_examples}
\end{table*}

Table \ref{tab:qualitative_examples} shows some examples of our conditional generation model's predictions.

In the first example, as the model has seen the mention ``Christmas truce'' with similar context during training, the model generates the exact title and description for the target entity. In fact, using this accurate profile, ES already ranks the target entity in the top 1 even without using the reranker.

In the second example, the model has not come across the mention ``Kevin Colbert'' during training. However, because of the phrases ``National Football League'' and ``general manager'', the model infers that the mention refers to an ``American football executive''. The generated description is quite close to the actual description, ``American football player and executive''. This generated profile helps ES rank the target entity higher than the entity Q91675515 (a researcher named Kevin Colbert).

The last example presents a failure case of our generation model. The target entity is a baseball team, but the model incorrectly infers that the mention ``Baltimore'' refers to a city. We will discuss this failure case in more detail in  next section. Nevertheless, if the hybrid approach is used, we can still recover from this error since the target entity is in the Wikipedia dictionary.
\subsection{Remaining Challenges} \label{sec:remaining_challenges}

In this section,  we will discuss some major categories of the remaining errors made by EPGEL.

\paragraph{Generation model's popularity bias} When encountering an input mention whose literal form has already appeared in the training set, the generation model sometimes ignores the context entirely and generates the most common entity profile for that literal form. In the last example in Table \ref{tab:qualitative_examples}, the mention Baltimore refers to a sports team. However, our model mistakenly generates the most common profile for the mention (a city in Maryland). A possible approach to tackle the challenge is to randomly mask out the input mention during training. This would encourage the generation model to pay more attention to the surrounding context and not rely too much on the mention's literal form.

\paragraph{Need to optimize global coherence} Entities within the same document are generally related; however, our reranker disambiguates each mention independently. Therefore, it sometimes makes mistakes that can be easily avoided if the global coherence among entities is considered. For example, given the following tweet, \textit{``\textcolor{red}{Syracuse} and \textcolor{red}{Pitt} in the \#ACC ... its gonna be a long year for \textcolor{red}{Maryland}.''}, EPGEL correctly infers that ``Syracuse'' and ``Pitt'' are basketball teams. However, for ``Maryland'', the reranker ranks a football team higher than the actual target entity (a basketball team). This shows that EPGEL may benefit from utilizing more global information for collective inference.
\section{Related Work}
\subsection{Candidate Entity Retrieval}
Dictionary-based techniques are the dominant approaches to candidate retrieval of many previous Wikipedia EL systems \cite{10.1145/2187836.2187898,10.14778/2536222.2536237,Shen2013LinkingNE,van2020rel}. The structure of Wikipedia provides a set of useful features for building an offline name dictionary between various names and their possible mapped entities. For example, many previous studies build such name dictionaries by mining anchor texts of Wikipedia pages \cite{Han2011CollectiveEL,Phan2017NeuPLAS,8354691}. Even though this approach is highly effective for EL over Wikipedia \cite{ganeahofmann2017deep}, it is not directly applicable to Wikidata as previously discussed.

\subsection{Entity Linking over Wikidata}
Compared to Wikipedia, there are relatively fewer studies on EL over Wikidata \cite{mollersurvey}. Recently, \newcite{Cetoli2019ANA} proposed a neural EL approach for Wikidata. The setting used in their work is that each mention comes with one correct entity candidate and one incorrect candidate. This setting is much less challenging and realistic than ours. \newcite{Sakor2020Falcon2A} proposed Falcon 2.0, a rule-based system for entity and relation linking over Wikidata. Its candidate retrieval approach is to query ES using the literal string of the input mention. This method is much less effective than our profiling-based approach (Sec. \ref{sec:cg_retrieval_evaluation}). OpenTapioca is another attempt that performs EL over Wikidata by utilizing two main features: local compatibility and semantic similarity \cite{Delpeuch2020OpenTapiocaLE}. For the social media domain, Tweeki \cite{harandizadehsingh2020tweeki} is an unsupervised approach for linking entities in tweets to Wikidata. EPGEL outperforms both OpenTapioca and Tweeki (Sec. \ref{sec:el_results_section}).
\section{Conclusions and Future Work}

This paper has proposed a novel profiling-based paradigm to candidate retrieval for EL. The technique is highly generalizable and complementary to the traditional dictionary-based approach, enabling the design of an effective hybrid candidate retrieval method. Together with a cross-attention reranker, our complete EL framework achieves strong performance on four public datasets. We plan to explore a broader range of properties and information about the target entity that can be extracted from the mention's context. For example, type-based features can be helpful for EL \cite{Onoe2020FineGrainedET}; as such, we aim to make our generation model generate the target entity's type. Also, in this work, we use a local model for candidate reranking. We plan to explore the use of a more global model for collective EL \cite{linetal2017list,yangetal2018collective,Phan2019PairLinkingFC}.

\section{Ethical Considerations}

\paragraph{Potential Risks} Our entity linking system has several potential malicious use cases (e.g., disinformation, generating fake news, surveillance). For example, \newcite{fungetal2021infosurgeon} introduced a novel approach for fake news generation. The technique works by first taking a genuine news article, extracting a multimedia knowledge graph, and replacing or inserting salient nodes or edges in the graph. To build such a multimedia knowledge graph, the authors do use an EL system. Another example is that our EL system may be used as part of a malicious surveillance system (e.g., automatically tracking the locations of celebrities based on social media posts and online news).

\paragraph{Limitations} Section \ref{sec:remaining_challenges} discusses some major categories of the remaining errors made by our entity linking system.

\section*{Acknowledgement}
We thank the anonymous reviewers for the helpful suggestions. 
This research is based upon work supported by U.S. DARPA KAIROS Program No. FA8750-19-2-1004 and U.S. DARPA AIDA Program No. FA8750-18-2-0014. The views and conclusions contained herein are those of the authors and should not be interpreted as necessarily representing the official policies, either expressed or implied, of DARPA, or the U.S. Government. The U.S. Government is authorized to reproduce and distribute reprints for governmental purposes notwithstanding any copyright annotation therein.

\bibliography{anthology,custom}
\bibliographystyle{acl_natbib}

\appendix
Section \ref{sec:appendix_eval_datasets} describes the datasets that we used for evaluation. Section \ref{sec:appendix_wikidata_preprocessing} describes how we preprocessed the original Wikidata dump. Section \ref{sec:appendix_reproducibility_checklist} presents our reproducibility checklist. Finally, Section \ref{sec:appendix_es_query_construction} describes how we construct an ES query from a generated profile.


\section{Evaluation Datasets}\label{sec:appendix_eval_datasets}
We use three different English datasets \cite{mollersurvey} for evaluating the performance of EL over Wikidata:

\begin{itemize}[topsep=0pt,itemsep=-1ex,partopsep=1ex,parsep=1ex]
\item \textbf{RSS-500} \cite{Rder2014NA} is a manually annotated dataset consisting of RSS-feeds (i.e., short formal documents) from major international newspapers. The target KB of the original version of RSS-500 is DBpedia. However, \newcite{Delpeuch2020OpenTapiocaLE} created a new version of the dataset for evaluating EL over Wikidata.
\item \textbf{ISTEX-1000} \cite{Delpeuch2020OpenTapiocaLE} is a dataset of 1,000 author affiliation strings extracted from scientific publications. It was manually annotated to align entity mentions to Wikidata.
\item \textbf{TweekiGold} \cite{harandizadehsingh2020tweeki} is a manually annotated dataset for EL over tweets. It has 500 tweets for evaluation but does not have a separate training set.
\end{itemize}

For RSS-500, ISTEX-1000, and WikipediaEL, the setting is that the gold-standard entity mentions are already given as input, and the task is only to link the input mentions to the correct entities.

For TweekiGold, similar to the study that introduced the dataset \cite{harandizadehsingh2020tweeki}, we do not assume that the mentions are provided. As such, for TweekiGold, we need to do both mention extraction and entity disambiguation. In this work, we simply use an off-the-shelf RoBERTa-based model from HuggingFace for mention extraction (roberta-base-finetuned-ner). Note that we do not fine-tune the mention extractor. In addition, when evaluating BLINK and GENRE on TweekiGold, we also use the same extractor to make the comparison fair.

For the TACKBP-2010 dataset \cite{ji2010overview}, there are 1,020 annotated mention/entity pairs in total for evaluation.  All the entities are from the TAC Reference KB, containing only 818,741 entities. However, to evaluate EPGEL, we use our large-scale Wikidata-based KB as the target KB.

RSS-500 and ISTEX-1000 can be downloaded from the Github repository of OpenTapioca \cite{Delpeuch2020OpenTapiocaLE}. And OpenTapioca is released under the Apache-2.0 license. TweekiGold is also released under the Apache-2.0 license. The TACKBP-2010 dataset can be downloaded from LDC's website. The license information can also be found at the LDC's website\footnote{\url{https://catalog.ldc.upenn.edu/LDC2018T16}}. Our use of the datasets is consistent with their licenses.

Our work focuses on English entity linking. In addition, we randomly sampled about 10$\sim$20 examples for each dataset and then checked whether the examples contained any offensive content. Overall, we did not see any example that had offensive content. 

\section{Wikidata Preprocessing}\label{sec:appendix_wikidata_preprocessing}
\begin{table}[!t]
\centering
\resizebox{0.95\linewidth}{!}{%
\begin{tabular}{|l|l|}
\hline
Wikidata ID & Label \\ \hline
Q4167836 & Wikimedia category \\ \hline
Q24046192 & Wikimedia category of stubs  \\ \hline
Q20010800 & Wikimedia user language category \\ \hline
Q11266439 & Wikimedia template\\ \hline
Q11753321 & Wikimedia navigational template \\ \hline
Q19842659 & Wikimedia user language template \\ \hline
Q21528878 & Wikimedia redirect page \\ \hline
Q17362920 & Wikimedia duplicated page \\ \hline
Q14204246 & Wikimedia project page \\ \hline
Q21025364 & WikiProject  \\ \hline
Q17442446 & Wikimedia internal item  \\ \hline
Q26267864 & Wikimedia KML file \\ \hline
Q4663903 & Wikimedia portal \\ \hline
Q15184295 & Wikimedia module \\ \hline
Q13442814 & Scholarly Article \\ \hline
\end{tabular}%
}
\caption{Wikidata identifiers used for filtering.}
\label{tab:filtering_ids}
\end{table}

In this work, we use the complete Wikidata dump dated August 2021. Even though Wikidata currently contains over 95 million items, many of the items are unhelpful (i.e., not entities we want to retain). Therefore, we apply several heuristics to filter out unuseful Wikidata items. First, we remove all entities with no English titles (i.e., entities whose English titles are empty strings). Second, we remove entities that are a subclass (P279) or instance of (P31) the most common Wikimedia-internal administrative entities. Table \ref{tab:filtering_ids} lists the Wikidata identifiers used for filtering (adapted from \cite{bothaetal2020entity,de2021multilingual}). Finally, we remove entities whose English titles start with ``Category:'', ``Template:'', or ``Project:''.

\section{Reproducibility Checklist}\label{sec:appendix_reproducibility_checklist} \label{sec:reproducibility_checklist}
In this section, we present the reproducibility information of the paper. We are planning to make the code publicly available after the paper is reviewed.

\paragraph{Implementation Dependencies Libraries} Pytorch 1.9.1 \cite{Paszke2019PyTorchAI}, Transformers 4.11.3 \cite{wolfetal2020transformers}, Numpy 1.19.5 \cite{Harris2020ArrayPW}, CUDA 11.2.

\paragraph{Computing Infrastructure} The experiments were conducted on a server with Intel(R) Xeon(R) Gold 5120 CPU @ 2.20GHz and NVIDIA Tesla V100 GPUs. Each GPU's memory is 16G.

\paragraph{Datasets} RSS-500 and ISTEX-1000 datasets can be downloaded from \url{https://github.com/wetneb/opentapioca}. The TweekiGold dataset can be downloaded from \url{https://ucinlp.github.io/tweeki/}. Finally, the TACKBP-2010 dataset can be downloaded from \url{catalog.ldc.upenn.edu/LDC2018T16}.

\paragraph{Number of Model Parameters} The number of parameters in the conditional generation model is about 140M. The number of parameters in the reranker is about 125M.

\paragraph{Hyperparameters} For training the conditional generation model, the batch size is set to be 128, the number of epochs is set to be 3, and the base learning rate is set to be 5e-5. For training the reranker, the batch size is set to be 8 mentions per batch (each mention has at most 50 candidates), the number of epochs is set to be 5, and the base learning rate is 1e-05.

\paragraph{Expected Validation Performance} The main paper has the results on the development set of WikipediaEL. We do not fine-tune our trained models on any of the evaluation datasets (i.e., RSS-500, ISTEX-1000, TweekiGold, and TACKBP-2010). For example, in Table \ref{tab:overall_el_results}, for EPGEL, we report the test results of the system with the best score on the development set of WikipediaEL.

\section{Elasticsearch Query Construction}\label{sec:appendix_es_query_construction}
\begin{figure*}[!ht]
  \centering
  \includegraphics[width=\textwidth]{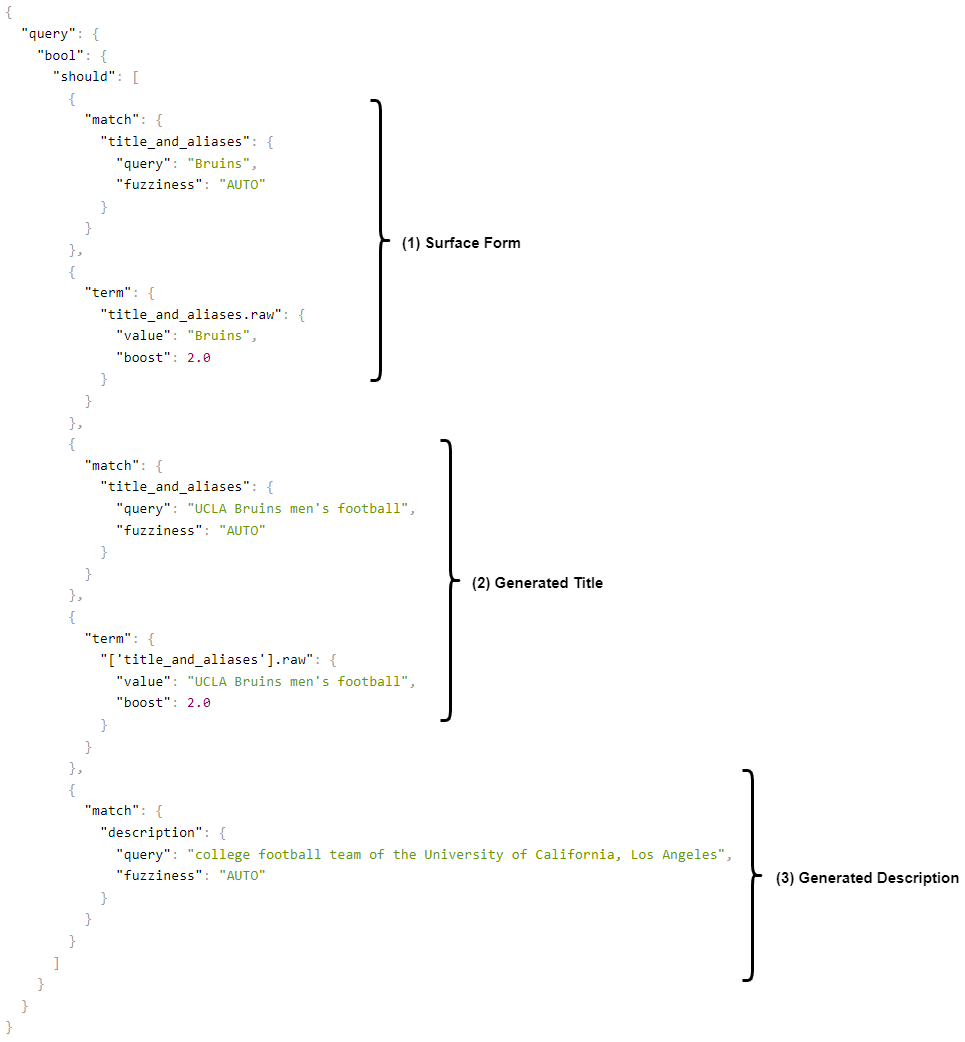}
  \caption{ES query for the example shown in Figure \ref{fig:candidate_generation}.}
  \label{fig:es_query}
\end{figure*}

We use the example shown in Figure \ref{fig:candidate_generation} as the running example. In this case, the surface form of the input mention is ``Bruins'', the generated title is ``UCLA Bruins men's football'', and the generated description is ``college football team of the University of California, Los Angeles''. Then, the actual query to be fed to ES is shown in Figure \ref{fig:es_query}. Intuitively, the query consists of three main parts:
\begin{enumerate}
    \item The similarity between the \textit{title} and \textit{alias} fields and the \textbf{surface form}.
    \item The similarity between the \textit{title} and \textit{alias} fields and the \textbf{generated title}.
    \item The similarity between the \textit{description} field and the \textbf{generated description}.
\end{enumerate}
Note that to reduce the querying latency, we merged the \textit{title} and \textit{alias} fields of each entity into one single field named \textit{title\_and\_aliases}. In other words, for each entity, its \textit{title\_and\_aliases} field is an array of strings corresponding to the entity's title and its aliases (if any). The \texttt{match} keyword is the standard keyword in ES for invoking a full-text search over a field. We use the \texttt{term} keyword to increase the final matching score when an exact match exists between the \textit{title\_and\_aliases} field and the surface form / the generated title. Overall, our ES query structure is quite basic and does not have many parameters. 

\end{document}